\pdfoutput=1

\documentclass[11pt]{article}
\usepackage[final]{acl}

\usepackage{times}
\usepackage{latexsym}
\usepackage[T1]{fontenc}
\usepackage[utf8]{inputenc}
\usepackage{microtype}
\usepackage{inconsolata}
\usepackage{graphicx}
\usepackage{multirow}

\usepackage{amsmath}
\usepackage{amssymb}
\usepackage{booktabs}
\usepackage{adjustbox}
\usepackage{float}

\title{Don’t Sweat the Small Stuff: Segment-Level Meta-Evaluation Based on Pairwise Difference Correlation}

\author{Colten DiIanni \and Daniel Deutsch \\
  Google \\
  \texttt{\{colediianni,dandeutsch\}@google.com} \\}

\begin{document}
\maketitle

\begin{abstract}
This paper introduces Pairwise Difference Pearson (PDP), a novel segment-level meta-evaluation metric for Machine Translation (MT) that address limitations in previous Pearson's $\rho$-based and and Kendall's $\tau$-based meta-evaluation approaches. PDP is a correlation-based metric that utilizes pairwise differences rather than raw scores. It draws on information from all segments for a more robust understanding of score distributions and uses segment-wise pairwise differences to refine Global Pearson to intra-segment score comparisons. Analysis on the WMT'24 shared task shows PDP properly ranks sentinel evaluation metrics and better aligns with human error weightings than previous work. Noise injection analysis demonstrates PDP's robustness to random noise, segment bias, and system bias while highlighting its sensitivity to extreme outliers.

\end{abstract}

\section{Introduction}
\label{sec:introduction}

Meta-evaluation of MT automatic metrics quantifies their performance using correlation between human-annotated scores ($Y$) with metric scores ($X$) for a set of translations \citep{WMT20}. The scores can be organized into $N \times M$ matrices, where $N$ is the number of evaluation systems and $M$ is the number of translations evaluated \citep{deutsch2023tiesmattermetaevaluatingmodern}. Alignment is often ranking-based, such as $acc_{eq}$ \citep[a derivative of Kendall's $\tau$ for handling tied scores;][]{deutsch2023tiesmattermetaevaluatingmodern} or correlation-based, such as Pearson's $\rho$.

Segment-level meta-evaluation assesses metric scores on individual translations, while system-level meta-evaluation measures system correlation or ranking agreement.
There are many ways to compute a segment-level agreement based on how scores are grouped together when calculating agreement and the specific agreement statistic that is used.
The two most common groupings either (1) calculate agreement using all values in $X$ and $Y$, denoted ``Global,'' or (2) calculate the average of $M$ correlations between each segment's $N$ translation scores, denoted ``Segment-Wise.''
Frequently used instantiations of these approaches that have been explored in the WMT Metrics Shared Task \citep{freitag-etal-2024-llms} are Global Pearson's $\rho$, Segment-Wise Pearson's $\rho$, and Segment-Wise $acc_{eq}$ (hereafter just $acc_{eq}$).

These three meta-evaluation metrics each have their own limitations.
Pearson correlations are sensitive to outliers \citep{mathur-etal-2020-tangled} and the segments analyzed under Segment-Wise Pearson may sample skewed score distributions due to small sample sizes. $acc_{eq}$ discards information about the magnitude of ranking differences.

This paper proposes Pairwise Difference Pearson (PDP), a novel meta-evaluation metric that addresses these limitations.
PDP computes a Pearson correlation on the pairwise differences between scores rather than the raw scores themselves and is able to draw on information from all segments for a more robust understanding of score distributions.
In this work, we define the properties of PDP, and present a comparative analysis against existing meta-evaluation statistics using MQM annotations from the WMT'23 and WMT'24 Metrics Shared Tasks \citep{WMT23,freitag-etal-2024-llms}. The difference between $acc_{eq}$ and PDP is then empirically tested using oracle metrics, showing PDP's effective correlation with human evaluation weights.

\section{Background and Related Work}
\label{sec:background}

Over the years, the methodology used by the WMT Metrics Shared Task has evolved and changed.
Below, we summarize the most commonly used methods for segment-level meta-evaluation.

\subsection{Kendall’s $\tau$ and $acc_{eq}$}
\label{sec:kendalls}

Kendall’s $\tau$ is a widely used ranking-based correlation coefficient in MT meta-evaluation \citep{WMT20,WMT21,WMT22,WMT23,freitag-etal-2024-llms}. It quantifies the proportion of agreement between metric and human rankings across all intra-segment translation pairs. Kendall's $\tau$ loses translation difference scale by looking only at rankings, meaning a small translation preference is equivalent to a large preference. This can lead to small errors biasing Kendall's $\tau$ as a meta-evaluation metric by flipping the preference between two similar translations.

The recent $acc_{eq}$ \citep{deutsch2023tiesmattermetaevaluatingmodern} refines traditional $\tau$ by adding correctly predicted ties to the set of concordant pairs. $acc_{eq}$ addresses the continuous nature of some metric predictions, where exact ties are rare. It does this by introducing a tie calibration procedure that broadens the definition of "ties" in the metric outputs.
Despite these changes, $acc_{eq}$ largely suffers from the same issues as $\tau$ due to it being a rank-based statistic.
Within WMT, $acc_{eq}$ has only been used Segment-Wise.

\subsection{Pearson at the Segment Level}
\label{sec:pearson}

Pearson’s $\rho$ measures the linear correlation between two vectors. In the context of segment-level meta-evaluation, Global Pearson flattens $X$ and $Y$ into vectors and calculates a single Pearson correlation.
For Segment-Wise Pearson, Pearson scores are computed for each individual segment and averaged to get the overall meta-evaluation score. 

In contrast to $acc_{eq}$, Pearson evaluates metrics considering the scale differences between vector values.
Appendix~\ref{appendix:pearson_proof} proves the equivalence between the Pearson correlation of a vector's raw values and the Pearson correlation on all the pairwise differences between the vector's raw values. This shows Segment-Wise Pearson is equivalent to Segment-Wise Pearson using pairwise differences between translations.

Segment-Wise Pearson solves the scale ignorance of $acc_{eq}$, but is extremely sensitive to noise. A limitation of Pearson's $\rho$ is its sensitivity to outliers \citep{mathur-etal-2020-tangled}. This issue is particularly pronounced in Segment-Wise Pearson's $\rho$ due to typically small input vectors ($N<30$).

The small sample size of segment vectors can cause misleading distributions, misleading the Pearson score. For example, a segment with all perfect translations except one with an insignificant error will rescale the minor error to an extreme outlier.

Global Pearson solves Segment-Wise Pearson's small sample size problem by calculating the Pearson score over all translations at once. While this approach better understands the overall human and metric score distributions, it introduces pairwise comparisons between translations from different segments. Considering the proof from Appendix~\ref{appendix:pearson_proof}, Global Pearson is equivalent to the Pearson correlation using pairwise differences between all translation scores in the dataset. This includes pairwise differences between translations from different source texts, which are not strictly comparable.
\section{Evaluating with PDP}
\label{sec:evaluating}

PDP is the Global Pearson correlation without direct inter-segment pairwise differences. The formula for PDP is outlined in Equation~\ref{eq:sgp}, where $X^*$ and $Y^*$ are $2(N^2) \times M$ matrices of the intra-segment pairwise differences of $X$ and $Y$. For each pair of translations ($x_1$, $x_2$), two pairwise differences are computed ($x_1$ - $x_2$, $x_2$ - $x_1$) to ensure the signs of $X^*$ and $Y^*$ values do not depend on the system ordering.

\begin{equation}
  \label{eq:sgp}
  \mathrm{PDP}(X, Y) = \mathrm{Global\ Pearson}(X^*, Y^*)
\end{equation}

PDP is different than Segment-Wise Pearson correlations because Segment-Wise Pearson's $\rho$ analyzes segments in isolation while PDP uses information from all segments at once, better understanding the overall score distribution. To distinguish PDP from Global Pearson, we consider the information loss introduced by PDP. Global Pearson calculates the correlation between all scores in $X$ and $Y$. This is equivalent to calculating the correlation between $X^{**}$ and $Y^{**}$, where $X^{**}$ and $Y^{**}$ are $2NM \times NM$ matrices of all score pairwise differences from $X$ and $Y$. Since PDP is calculated using intra-segment pairwise differences $X^*$ and $Y^*$, it effectively removes the raw score pairwise differences between segments. This information loss targets PDP towards intra-segment differences rather than score correlation across segments.

Since PDP uses Pearson correlation as the underlying metric, it is sensitive to outliers (\S\ref{sec:pearson}). However, PDP does not suffer from the "NaN problem", as detailed by \citet{deutsch2023tiesmattermetaevaluatingmodern}. If all pairwise difference predictions are constant, resulting in an undefined Pearson's $\rho$, we assign a value of 0, indicating no correlation with ground truth scores. This constant scoring scenario under PDP is less common than under Segment-Wise Pearson's $\rho$ because PDP considers all $N \times M$ scores while segment-wise Pearson's $\rho$ is computed using $N$ scores at a time.
\section{Analysis Setup}
\label{sec:setup}

\subsection{Datasets}

For empirical meta-evaluation, we used the Multidimensional Quality Metrics \citep[MQM; ][]{Lommel14,Freitag21} annotations provided by the WMT'23 \citep{freitag-etal-2023-results} and WMT'24 \citep{freitag-etal-2024-llms} metrics shared tasks. The MQM scores serve as the ground-truth against which automatic metrics were evaluated. 

On the WMT'23 metrics shared task, our analysis encompasses two language pairs: English to German (en$\rightarrow$de) and Chinese to English (zh$\rightarrow$en). The datasets for these language pairs contain 12-15 MT evaluation systems and 557-1976 segments. WMT'23 also includes two additional rounds of human annotations to measure inter-annotator agreement. On the WMT'24 metrics shared task, our analysis uses three language pairs: English to German (en$\rightarrow$de), Japanese to Chinese (ja$\rightarrow$zh), and English to Spanish (en$\rightarrow$es), with an emphasis on the en$\rightarrow$de language pair. The datasets for these language pairs contain 21 to 26 MT evaluation systems and 722 to 998 segments.

\subsection{Automatic Metrics Under Evaluation}

We conducted segment-level meta-evaluation on the set of automatic metrics submitted to the WMT'23 and WMT'24 shared tasks. The WMT'24 shared task includes three sentinel metrics \citep{perrella-etal-2024-guardians}, designed to assess the fairness characteristics of each meta-evaluation metric. These sentinel metrics were trained using MQM and Direct Assessment (DA) data from WMT 2017-2022 \citep{WMT17, WMT18, WMT19, WMT20, WMT21, WMT22}, with each being provided specific, limited information about the translation during evaluation:

\begin{itemize}
    \item \texttt{sentinel-cand-mqm}: Score translations based only on the candidate translation.
    \item \texttt{sentinel-src-mqm}: Score translations based only on the source text.
    \item \texttt{sentinel-ref-mqm}: Score translations based only on the reference translation.
\end{itemize}

While the \texttt{src} and \texttt{ref} sentinels consistently produce a constant score across all translations within a given segment, \texttt{sentinel-cand-mqm}'s score can vary within a single segment, rendering \texttt{sentinel-cand-mqm} a valuable benchmark for segment-level meta-evaluation. We hypothesize this metric primarily captures fluency and stylistic errors rather than accuracy errors. As such, we expect it should be patently outperformed by SOTA evaluation metrics. Since the true ranking of metrics is unknown, it is not possible to definitively say which meta-evaluation metric is better. Therefore we focus on the sentinel metrics, which should be ranked low, and agreement with other segment-level meta-evaluation metrics.
\section{Analysis}
\label{sec:analysis}

\begin{table}[]
\begin{adjustbox}{width=\columnwidth}
\begin{tabular}{lrrrr}
\toprule
\multirow{2}{*}{\bf Metric} & \multirow{2}{2.4cm}{\centering \textbf{Segment-Wise\\Pearson}} & \multirow{2}{2cm}{\centering \textbf{Global Pearson}} & \multirow{2}{*}{\centering $\mathbf{acc_{eq}}$} & \multirow{2}{*}{\centering $\mathbf{PDP}$} \\ \\
\midrule
XCOMET & 0.404 (\phantom{1}2) & 0.459 (\phantom{1}1) & 0.530 (\phantom{1}3) & 0.443 (\phantom{1}1) \\
XCOMET-QE* & 0.355 (\phantom{1}8) & 0.428 (\phantom{1}3) & 0.520 (\phantom{1}5) & 0.397 (\phantom{1}2) \\
metametrics & 0.419 (\phantom{1}1) & 0.437 (\phantom{1}2) & 0.542 (\phantom{1}1) & 0.393 (\phantom{1}3) \\
MetricX-24-Hybrid & 0.403 (\phantom{1}3) & 0.393 (\phantom{1}4) & 0.532 (\phantom{1}2) & 0.383 (\phantom{1}4) \\
bright-qe* & 0.261 (14) & 0.353 (\phantom{1}6) & 0.500 (\phantom{1}8) & 0.350 (\phantom{1}5) \\
MetricX-24-Hybrid-QE* & 0.379 (\phantom{1}5) & 0.336 (\phantom{1}7) & 0.526 (\phantom{1}4) & 0.349 (\phantom{1}6) \\
COMET-22 & 0.381 (\phantom{1}4) & 0.311 (\phantom{1}9) & 0.482 (11) & 0.322 (\phantom{1}7) \\
metametrics\_qe* & 0.221 (19) & 0.357 (\phantom{1}5) & 0.497 (\phantom{1}9) & 0.321 (\phantom{1}8) \\
gemba\_esa* & 0.361 (\phantom{1}7) & 0.282 (12) & 0.507 (\phantom{1}7) & 0.319 (\phantom{1}9) \\
BLCOM\_1 & 0.350 (\phantom{1}9) & 0.283 (11) & 0.455 (13) & 0.300 (10) \\
CometKiwi* & 0.244 (17) & 0.229 (16) & 0.467 (12) & 0.288 (11) \\
MEE4 & 0.257 (15) & 0.190 (19) & 0.437 (16) & 0.282 (12) \\
BLEURT-20 & 0.372 (\phantom{1}6) & 0.332 (\phantom{1}8) & 0.486 (10) & 0.281 (13) \\
chrfS & 0.246 (16) & 0.153 (21) & 0.434 (20) & 0.253 (14) \\
BERTScore & 0.229 (18) & 0.201 (18) & 0.435 (18) & 0.241 (15) \\
\texttt{sentinel-cand-mqm}* & 0.298 (11) & 0.306 (10) & 0.517 (\phantom{1}6) & 0.223 (16) \\
PrismRefMedium & 0.307 (10) & 0.146 (23) & 0.434 (19) & 0.222 (17) \\
PrismRefSmall & 0.295 (12) & 0.142 (24) & 0.433 (21) & 0.212 (18) \\
damonmonli & 0.178 (23) & 0.261 (14) & 0.443 (15) & 0.210 (19) \\
YiSi-1 & 0.282 (13) & 0.202 (17) & 0.436 (17) & 0.208 (20) \\
chrF & 0.220 (20) & 0.142 (25) & 0.431 (22) & 0.192 (21) \\
spBLEU & 0.216 (21) & 0.155 (20) & 0.431 (24) & 0.161 (22) \\
BLEU & 0.196 (22) & 0.149 (22) & 0.431 (23) & 0.151 (23) \\
XLsimMqm* & 0.087 (24) & 0.080 (26) & 0.450 (14) & 0.005 (24) \\
\texttt{sentinel-ref-mqm} & 0.000 (25) & 0.246 (15) & 0.429 (25) & 0.000 (25) \\
\texttt{sentinel-src-mqm}* & 0.000 (25) & 0.262 (13) & 0.429 (25) & 0.000 (25) \\
\bottomrule
\end{tabular}
\end{adjustbox}
\caption{\label{wmt24_en-de_metric_comparison_table}
Scores (and ranks) of metrics evaluated by Segment-Wise Pearson, Global Pearson, $acc_{eq}$, and PDP on the WMT’24 en$\rightarrow$de dataset. QE metrics are marked with a *.}
\end{table}

Table~\ref{wmt24_en-de_metric_comparison_table} presents a comparative analysis of segment-level performance under Segment-Wise Pearson's $\rho$, Global Pearson's $\rho$, $acc_{eq}$, and PDP for the en$\rightarrow$de language pair of the WMT'24 metrics shared task. Segment-Wise Pearson rankings disagree with many other meta-evaluation metrics, ranking XCOMET-QE $8^{th}$, bright-qe $14^{th}$, and metametrics\_qe $19^{th}$. Global Pearson ranks the \texttt{sentinel-src} and \texttt{sentinel-ref} $13^{th}$ and $15^{th}$ despite each system predicting only ties within every segment. This shows how Global Pearson uses inter-segment correlations for meta-evaluation.

A key difference between PDP and all other segment-level meta-evaluation rankings is for \texttt{sentinel-cand}. While other meta-evaluation metrics rank \texttt{sentinel-cand} at $11^{th}$ and above, PDP ranks it $16^{th}$ out of 26. The divergent ranking of \texttt{sentinel-cand} between PDP and $acc_{eq}$ is also observed using zh$\rightarrow$en data in Appendix~\ref{appendix:other_lps}.

Under the assumption that humans are better raters of MT quality than automatic metrics, a meta-evaluation metric should rank human re-annotations highest, showing it is a reliable measure of correlation with human judgment. Existing meta-eval metrics failing to do this is a problem that has been recently demonstrated by \citet{proietti2025machinetranslationevaluationachieved}. The inter-annotator agreement was first calculated under $acc_{eq}$ and PDP on the WMT'23 shared task. PDP ranks the second and third rounds of human annotation $\mathrm{1}^{st}$ and $\mathrm{2}^{nd}$ under en$\rightarrow$de, while $acc_{eq}$ ranks them $\mathrm{5}^{th}$ and $\mathrm{8}^{th}$. Human annotations also rank higher using PDP than $acc_{eq}$ under zh$\rightarrow$en. Details are provided in Appendix~\ref{appendix:wmt_23_analysis}.

\subsection{Oracle Metrics}
\label{sec:oracle_sentinels}

To investigate \texttt{sentinel-cand}'s meta-metric ranking difference between $acc_{eq}$ and PDP, we consider how $acc_{eq}$ and PDP show bias towards different error categories. We believe good meta-evaluation metrics should rank accuracy-focused metrics above fluency-focused ones, as fluency errors often preserve source text meaning.

An oracle metric was constructed for each MQM error category by aggregating all the error category's MQM errors and was evaluated against the total human error scores. These oracle sentinels, as ideal detectors for their error types, allow direct examination of how meta-evaluation metrics weigh each error's importance. 

We define an error category's importance as its total contribution to the human scores: the sum of all error category annotations across the MQM dataset. Full information about each error category's importance, count, and weight are given in Appendix~\ref{appendix:error_weights_table}. For each error category's corresponding oracle metric, we find its predictive power under $acc_{eq}$ and PDP. A Spearman rank correlation is used to measure correlation between each metrics' importance and meta-evaluation score. $acc_{eq}$ and PDP perform similarly well, with Spearman correlations of 0.90 and 0.88 respectively.

The two factors which determine an error category's importance are the number of errors and the average error weight. We believe error weight reflects the value human annotators place on each error category, measuring a single error's effect on the score. Oracle metric PDP scores are better correlated with their respective error weights, with PDP achieving Spearman correlation of 0.74 and $acc_{eq}$ only 0.30. $acc_{eq}$ is more correlated with the error counts than PDP, achieving 0.66 and 0.40 respectively. 

These results highlight a key distinction: PDP emphasizes error weight and is less sensitive to many small errors than $acc_{eq}$. $acc_{eq}$'s sensitivity to error count is an attribute of its binary view of pairwise differences. Small score differences, particularly in the human scores, can disproportionately impact $acc_{eq}$ if they change the translation rankings. This analysis provides an explanation for why \texttt{sentinel-cand} ranks higher under $acc_{eq}$ than PDP: the oracle may correctly identify many fluency-based errors, but such errors are not heavily weighted by human annotators. 
\section{Robustness to Noise}
\label{06_noise_robustness}

Selecting a meta-evaluation metric is challenging due to the lack of a ground truth ranking for evaluation metrics. We introduce synthetic datasets with artificially injected noise to measure the effect of perturbations on meta-evaluation metric scores. For our ground truth dataset ($Y$), we chose the en$\rightarrow$de MQM scores. $Y$ contains negative values ranging from -100 to 0, with individual errors contributing -1 for minor errors and -5 for major errors, as determined by raters \citep{freitag-etal-2021-experts}. Most scores are between -25 and 0, with non-translations scoring -25. From the ground truth, the best and worst scores were calculated for each meta-evaluation metric by comparing the MQM scores against themselves and against randomly guessed scores from the distribution of $Y$, ($X_{rand}$).

\begin{figure*}[!t]
    \centering
    \begin{adjustbox}{width=\textwidth, height=150px, keepaspectratio}
    \includegraphics[width=\columnwidth]{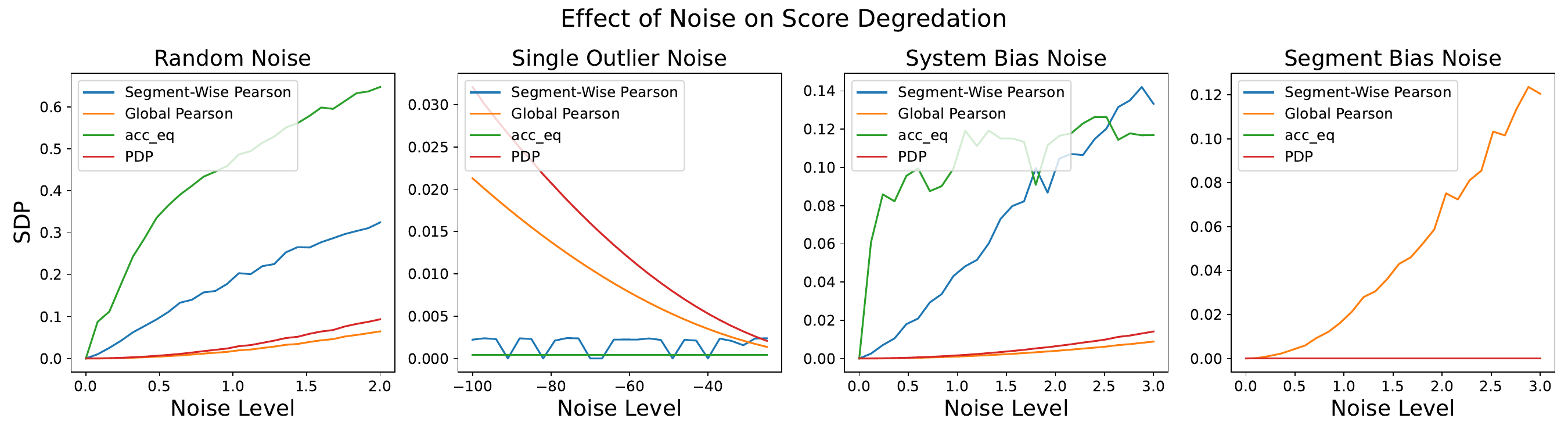}
    \end{adjustbox}
    \caption{SDP for Segment-Wise Pearson's $\rho$, Global Pearson's $\rho$, $acc_{eq}$, and PDP under increasing levels of noise. Lower SDP values indicate greater stability and robustness to noise.}
    \label{fig:noise_robustness}
\end{figure*}

To quantify the effect of artificial noise added to $Y$, each meta-evaluation metric's score degradation was measured under increasing levels of noise. The formulation of score degradation proportion (SDP) measures the meta-evaluation metric's ($\theta$) score change from ground truth scaled by the theoretical range of score change (Equation~\ref{eq:score_degradation}). $X_{noise}$ is the $N \times M$ matrix $Y$ with noise added to it and $X_{rand}$ is a $N \times M$ matrix of score predictions generated by randomly sampling from $Y$ with replacement.

\begin{equation}
  \label{eq:score_degradation}
  \mathrm{SDP}(Y|\theta) = \frac{\theta(Y, Y) - \theta(Y, X_{noise})}{\theta(Y, Y) - \theta(Y, X_{rand})}
\end{equation}

SDP indicates a meta-evaluation metric's noise sensitivity; better metrics are expected to degrade less given noisy versions of ground truth predictions. Four variants of noise injections, each testing unique aspects of robustness, were tested:

\begin{itemize}
    \item random noise: for each element of $Y$, add a random value sampled from ${\sim} N(0,noise)$
    \item extreme outlier: a randomly selected element of $Y$ is set to $noise$
    \item system bias: for a randomly selected system of $Y$, add $noise$ to all scores of the system
    \item segment bias: for each segment of $Y$, add a random value sampled from ${\sim} N(0,noise)$
\end{itemize}

Figure~\ref{fig:noise_robustness} visualizes the performance of Segment-Wise Pearson, Global Pearson, $acc_{eq}$, and PDP under varying levels of noisy conditions. The leftmost plot in Figure~\ref{fig:noise_robustness} illustrates the effect of increasing levels of random noise injection on meta-evaluation metric SDP. As the random noise level increases, Global Pearson and PDP exhibit the most robust performance. Their consistently low SDP indicates these meta-evaluation scores are least affected by random noise in the evaluation data.

While Global Pearson and PDP are more robust to random noise, they are less robust to a single, extreme outlier. Since the segments are analyzed in isolation using Segment-Wise Pearson and $acc_{eq}$ does not consider scale, these two meta-evaluation metrics cap the effect of individual outliers, thereby providing greater robustness. While PDP is less robust to a single extreme outlier, we believe this sensitivity is less concerning than sensitivity to random noise. Random noise is assumed to be an inherent part of the data, whereas outliers can often be identified through data inspection and managed using techniques such as score clipping.

Ideally, segment-level meta-evaluation metrics would be less sensitive to system bias, as system-level meta-evaluation metrics are designed to capture this. The third plot in Figure~\ref{fig:noise_robustness} simulates unfairly biasing a metric towards a single system. Global Pearson and PDP are more robust to system bias than the other metrics tested.

The final plot in Figure~\ref{fig:noise_robustness} simulates an evaluation metric which is biased by the source text, with the relative rankings within each segment remaining unchanged. Global Pearson is the only metric affected by this noise. This confirms the findings in Section~\ref{sec:analysis}: Global Pearson is not restricted to intra-segment comparisons while PDP is.
\section{Conclusion}
\label{sec:conclusion}

This work introduces PDP for MT segment-level meta-evaluation. PDP addresses limitations in metrics like Segment-Wise Pearson's $\rho$, Global Pearson's $\rho$, and $acc_{eq}$ by using pairwise score differences from multiple segments for more accurate distribution estimation. Meta-evaluation on the WMT'23 shared task ranked human evaluation higher under PDP than $acc_{eq}$. WMT'24 shared task analysis showed PDP consistently outperforms other segment-level meta-evaluation metrics, down-ranking sentinel metrics and better aligning with human error weightings. While our analysis focuses on MT, PDP is generalizable to any segment-level meta-evaluation task in NLP.
\section{Limitations}
\label{sec:limitations}

Our analysis is limited to four language pairs from the WMT'23 and WMT'24 metrics shared tasks. Section~\ref{06_noise_robustness} details PDP's sensitivity to outliers. This is a known limitation of Pearson's $\rho$ as a correlation-based statistic. Although PDP's random noise robustness may overshadow its outlier sensitivity, this tradeoff will depend on the use case and dataset. PDP also assumes a consistent scoring variance between raters.

\bibliography{custom}

\appendix

\section{Pearson Correlation Using Pairwise Differences}
\label{appendix:pearson_proof}

In this section, we prove the Pearson correlation between two vectors is equivalent to the Pearson correlation between the pairwise difference counterparts of each vector.

\subsection{Pearson Equivalence Proof}
\label{appendix:proof}

\begin{flushleft}
Let $X = (x_1, x_2, ..., x_n)$ and $Y = (y_1, y_2, ... y_n)$. The vector of all pairwise differences for $X$ is defined:

\begin{equation*}
\begin{split}
\Delta X =
& (x_1 - x_1, x_1 - x_2, \dots, x_1 - x_n, \\
& x_2 - x_1, x_2 - x_2, \dots, x_2 - x_n, \\ 
& \dots, x_n - x_1, \dots, x_n - x_n)
\end{split}
\end{equation*}

This vector $\Delta X$ has $N = n^2$ elements. We denote elements of $\Delta X$ as $(\Delta X)_k$ where $k$ indexes a pair $(i, j)$, so $(\Delta X)_k = x_i - x_j$.

Similarly for $Y$:
\begin{equation*}
\begin{split}
\Delta Y =
& (y_1 - y_1, y_1 - y_2, \dots, y_1 - y_n, \\
& y_2 - y_1, y_2 - y_2, \dots, y_2 - y_n, \\
& \dots, y_n - y_1, \dots, y_n - y_n)
\end{split}
\end{equation*}

Each element $(\Delta Y)_k = y_i - y_j$ corresponds to the same pair of indices $(i, j)$ as for $(\Delta X)_k$.

The Pearson correlation coefficient between two vectors $U$ and $V$ is $P(U, V) = \frac{Cov(U, V)}{\sqrt{Var(U)Var(V)}}$.

We want to prove $P(X, Y) = P(\Delta X, \Delta Y)$

1. \textbf{Mean of $\Delta X$ and $\Delta Y$}

The mean of $\Delta X$ is:
\end{flushleft}

\begin{equation}
\label{eq:mean_of_delta_x}
\begin{aligned}
\Delta \bar{X}
&= \frac{1}{n^2}\Sigma_{i=1}^n\Sigma_{j=1}^n (x_i - x_j)\\
&= \frac{1}{n^2}(\Sigma_{i=1}^n\Sigma_{j=1}^n x_i - \Sigma_{i=1}^n\Sigma_{j=1}^n x_j)\\
&= \frac{1}{n^2}(n\Sigma_{i=1}^n x_i - n\Sigma_{j=1}^n x_j)\\
&= \bar{X} - \bar{X}\\
&= 0
\end{aligned}
\end{equation}

\begin{flushleft}
So, $\Delta \bar{X} = 0$. Similarly, $\Delta \bar{Y} = 0$.

2. \textbf{Variance of $\Delta X$ and $\Delta Y$}

Since $\Delta \bar{X} = 0$ (Equation~\ref{eq:mean_of_delta_x}), we can show:
\end{flushleft}
\begin{equation}
\label{eq:variance_of_delta_x}
\begin{aligned}
&\text{Var}(\Delta X)\\
&= \mathbb{E}[(\Delta X)^2]\\
&= \frac{1}{n^2}\Sigma_{i=1}^n\Sigma_{j=1}^n (x_i - x_j)^2\\
&= \frac{1}{n^2}\Sigma_{i=1}^n\Sigma_{j=1}^n ((x_i - \bar{X}) - (x_j - \bar{X}))^2\\
& \text{Let}\ x'_k = x_k - \bar{X}.\\
&= \frac{1}{n^2}\Sigma_{i=1}^n\Sigma_{j=1}^n (x'_i - x'_j)^2\\
&= \frac{1}{n^2}\Sigma_{i=1}^n\Sigma_{j=1}^n ((x'_i)^2 - 2x'_ix'_j + (x'_j)^2)\\
&= \frac{1}{n}\Sigma_{i=1}^n(x'_i)^2 - \frac{2}{n^2}\Sigma_{i=1}^n x'_i\Sigma_{j=1}^nx'_j\\
& \quad + \frac{1}{n}\Sigma_{j=1}^n(x'_j)^2\\
& \text{Since}\ \Sigma_{k=1}^n x'_k  = 0:\\
&= \frac{1}{n}\Sigma_{i=1}^n(x'_i)^2 + \frac{1}{n}\Sigma_{j=1}^n(x'_j)^2\\
&= \frac{2}{n}\Sigma_k^n (x'_k)^2\\
&= \frac{2}{n}(n \times \text{Var}(X))\\
&= 2 \times \text{Var}(X)\\
\end{aligned}
\end{equation}
Similarly, $\text{Var}(\Delta Y) = 2 \times \text{Var}(Y)$.

\begin{flushleft}

3. \textbf{Covariance of $\Delta X$ and $\Delta Y$}

Since $\Delta \bar{X} = 0$ and $\Delta \bar{Y} = 0$: 
\end{flushleft}
\begin{equation}
\label{eq:covariance_of_delta_x_and_delta_y}
\begin{aligned}
& \text{Cov}(\Delta \bar{X}, \Delta \bar{Y})\\
&= \mathbb{E}[\Delta \bar{X}\Delta \bar{Y}]\\
&= \frac{1}{n^2}\Sigma_{i=1}^n\Sigma_{j=1}^n (x_i - x_j)(y_i - y_j)\\
& \text{Let}\ x'_k = x_k - \bar{X}\ \text{and}\ y'_k = y_k - \bar{Y}.\\
&= \frac{1}{n^2}\Sigma_{i=1}^n\Sigma_{j=1}^n (x'_i - x'_j)(y'_i - y'_j)\\
&= \frac{1}{n^2}\Sigma_{i=1}^n\Sigma_{j=1}^n (x'_iy'_i - x'_jy'_i\\
& \quad - x'_iy'_j + x'_jy'_j)\\
&= \frac{1}{n}\Sigma_{i=1}^nx'_iy'_i + \frac{1}{n}\Sigma_{j=1}^nx'_jy'_j\\
&= \frac{2}{n}\Sigma_{k=1}^nx'_ky'_k\\
&= \frac{2}{n}(n \times \text{Cov}(X, Y))\\
&= 2 \times \text{Cov}(X, Y)\\
\end{aligned}
\end{equation}

\begin{flushleft}
So, $\text{Cov}(\Delta \bar{X}, \Delta \bar{Y}) = 2 \times \text{Cov}(X, Y)$.

4. \textbf{Correlation of $\Delta X$ and $\Delta Y$}
\end{flushleft}

\begin{equation}
\label{eq:correlation_of_delta_x_and_delta_y}
\begin{aligned}
P(\Delta X, \Delta Y)
&= \frac{\text{Cov}(\Delta X, \Delta Y)}{\sqrt{\text{Var}(\Delta X)Var(\Delta Y)}}\\
&= \frac{2\text{Cov}(X, Y)}{\sqrt{2\text{Var}(X) \times 2\text{Var}(Y)}}\\
&= \frac{\text{Cov}(X, Y)}{\sqrt{\text{Var}(X)\text{Var}(Y)}}\\
&= P(X, Y)\\
\end{aligned}
\end{equation}

\begin{flushleft}
This equality holds assuming $\text{Var}(X) > 0$ and $\text{Var}(Y) > 0$. If either variance is zero, both correlations are undefined.
\end{flushleft}

\section{Other Language Pair Analysis}
\label{appendix:other_lps}

Table~\ref{ja-zh_metric_comparison_table} and Table~\ref{en-es_metric_comparison_table} present the segment-level performance of the zh$\rightarrow$en and en$\rightarrow$es language pairs. When using PDP instead of $acc_{eq}$ the \texttt{sentinel-cand} ranking falls from $12^{th}$ to $21^{st}$ using zh$\rightarrow$en and raised from $10^{th}$ to $7^{th}$ using en$\rightarrow$es. We believe the ranking change under en$\rightarrow$es is less reliable because many of the translations for this dataset received perfect human scores, resulting in many segment-wise ties. We see the en$\rightarrow$es high tie rate's effect reflected in the smaller $acc_{eq}$ score range, with the worst and best scoring systems achieving $acc_{eq}$ performances of 0.680 and 0.689 respectively.
\subsection{WMT'23}
\label{appendix:wmt_23_analysis}

The segment-level performance under Segment-Wise Pearson's $\rho$, Global Pearson's $\rho$, $acc_{eq}$, and PDP for the WMT'23 en$\rightarrow$de language pair is shown in Tables~\ref{wmt23_en-de_metric_comparison_table} and ~\ref{wmt23_zh-en_metric_comparison_table}. These tables include two rounds of human reratings (human-round2 and human-round3) which are ranked highest in total undern PDP than all other segment-level meta-evaluation metrics.

\section{PDP vs. $acc_{eq}$: Error Weight}
\label{appendix:error_weights_table}

Using the WMT'24 metrics shared task en$\rightarrow$de dataset, sentinel metrics were constructed by filtering MQM annotations for each error category. These sentinels simulate a perfect evaluation model for their respective error types. Table~\ref{en-de_error_weights_table} details the number and average severity of annotations labeled by each sentinel. The product of the count and average weight is a measure of the overall weight each error category contributes to the final evaluation: importance. Using these sentinel metrics, we can analyze how each meta-evaluation metric values different types of errors in Section~\ref{sec:oracle_sentinels}.

\begin{table}[]
    \centering
    \begin{adjustbox}{width=\columnwidth}
    \begin{tabular}{l|r|rr|rr}
        \toprule
        \textbf{Error Category} & \textbf{Importance} & \textbf{Count} & \textbf{Avg. Weight} & $\mathbf{acc_{eq}}$ & $\mathbf{PDP}$ \\
        \midrule
        accuracy/addition & \phantom{2}925 (\phantom{1}9) & 249 (13) & 3.715 (\phantom{1}6) & 0.451 (\phantom{1}9) & 0.310 (\phantom{1}4) \\
        accuracy/creative reinterpretation & \phantom{4}0 (22) & 770 (\phantom{1}5) & 0.000 (22) & 0.429 (21) & 0.000 (21) \\
        accuracy/gender mismatch & \phantom{2}156 (16) & 32 (21) & 4.875 (\phantom{1}2) & 0.433 (15) & 0.105 (13) \\
        accuracy/mistranslation & 12749 (\phantom{1}1) & 3805 (\phantom{1}1) & 3.351 (\phantom{1}9) & 0.646 (\phantom{1}1) & 0.600 (\phantom{1}1) \\
        accuracy/omission & \phantom{2} 767 (10) & 187 (15) & 4.102 (\phantom{1}4) & 0.450 (10) & 0.320 (\phantom{1}3)\\
        accuracy/source language fragment & \phantom{1}1690 (\phantom{1}4) & 438 (\phantom{1}8) & 3.858 (\phantom{1}5) & 0.464 (\phantom{1}6) & 0.218 (\phantom{1}7) \\
        fluency/grammar & \phantom{1}2088 (\phantom{1}3) & 1052 (\phantom{1}4) & 1.985 (11) & 0.513 (\phantom{1}4) & 0.265 (\phantom{1}6) \\
        fluency/inconsistency & \phantom{2}369 (12) & 137 (17) & 2.693 (\phantom{1}8) & 0.442 (12) & 1.109 (12) \\
        fluency/punctuation & \phantom{2}262 (14) & 1884 (\phantom{1}2) & 0.139 (21) & 0.526 (\phantom{1}3) & 0.038 (15) \\
        fluency/register & \phantom{1}1491 (\phantom{1}5) & 415 (\phantom{1}9) & 3.593 (\phantom{1}7) & 0.461 (\phantom{1}7) & 0.271 (\phantom{1}5) \\
        fluency/spelling & \phantom{2}978 (\phantom{1}7) & 658 (\phantom{1}6) & 1.486 (13) & 0.478 (\phantom{1}5) & 0.123 (11) \\
        fluency/text-breaking & \phantom{2}192 (15) & 108 (19) & 1.778 (12) & 0.437 (13) & 0.031 (16) \\
        locale convention/address format & \phantom{4}2 (21) & 2 (23) & 1.000 (17) & 0.429 (21) & 0.007 (20) \\
        locale convention/currency format & \phantom{4}7 (20) & 3 (22) & 2.333 (10) & 0.430 (18) & 0.017 (18) \\
        locale convention/time format & \phantom{4}8 (19) & 8 (20) & 1.000 (18) & 0.430 (18) & 0.018 (17) \\
        non-translation! & \phantom{2}975 (\phantom{1}8) & 39 (20) & 25.000 (\phantom{1}1) & 0.431 (16) & 0.459 (\phantom{1}2) \\
        other & \phantom{2}267 (13) & 87 (19) & 3.069 (10) & 0.437 (13) & 0.070 (14) \\
        source issue & \phantom{4}0 (22) & 1791 (\phantom{1}3) & 0.000 (22) & 0.429 (21) & 0.000 (21) \\
        style/archaic or obscure word choice & \phantom{3}42 (17) & 14 (24) & 3.000 (12) & 0.430 (18) & 0.011 (19) \\
        style/bad sentence structure & \phantom{2}521 (11) & 173 (16) & 3.012 (11) & 0.450 (10) & 0.197 (\phantom{1}9) \\
        style/unnatural or awkward & \phantom{1}3620 (\phantom{1}2) & 1996 (\phantom{1}2) & 1.814 (14) & 0.558 (\phantom{1}2) & 0.201 (\phantom{1}8) \\
        terminology/inappropriate for context & \phantom{1}1021 (\phantom{1}6) & 349 (10) & 2.926 (\phantom{1}7) & 0.460 (\phantom{1}8) & 0.156 (10) \\
        terminology/inconsistent & \phantom{3}23 (18) & 19 (19) & 1.211 (16) & 0.431 (16) & -0.004 (23) \\
        \bottomrule
    \end{tabular}
    \end{adjustbox}
    \caption{\label{en-de_error_weights_table}
        Importance, count, and avg. weight (with rank) for each error category in the WMT'24 en$\rightarrow$de human evaluations. $acc_{eq}$ and PDP scores (with ranks) are included for oracle sentinel metrics which score translations based only the category's ground truth MQM errors.}
\end{table}

\clearpage

\begin{table}[]
\begin{adjustbox}{width=\columnwidth}
\begin{tabular}{lrrrr}
\toprule
\textbf{Metric} & \textbf{Segment-Wise Pearson} & \textbf{Global Pearson} & $\mathbf{acc_{eq}}$ & $\mathbf{PDP}$\\
\midrule
metametrics & 0.413 (\phantom{1}1) & 0.475 (\phantom{1}1) & 0.561 (\phantom{1}1) & 0.408 (\phantom{1}1) \\
MetricX-24-Hybrid-QE* & 0.360 (\phantom{1}7) & 0.432 (\phantom{1}2) & 0.530 (\phantom{1}4) & 0.407 (\phantom{1}2) \\
MetricX-24-Hybrid & 0.398 (\phantom{1}3) & 0.430 (\phantom{1}3) & 0.539 (\phantom{1}3) & 0.400 (\phantom{1}3) \\
gemba\_esa* & 0.405 (\phantom{1}2) & 0.400 (\phantom{1}6) & 0.539 (\phantom{1}2) & 0.356 (\phantom{1}4) \\
XCOMET & 0.395 (\phantom{1}4) & 0.427 (\phantom{1}4) & 0.510 (\phantom{1}6) & 0.346 (\phantom{1}5) \\
COMET-22 & 0.383 (\phantom{1}5) & 0.366 (\phantom{1}9) & 0.496 (\phantom{1}7) & 0.340 (\phantom{1}6) \\
metametrics\_qe* & 0.281 (16) & 0.405 (\phantom{1}5) & 0.516 (\phantom{1}5) & 0.314 (\phantom{1}7) \\
damonmonli & 0.273 (18) & 0.337 (11) & 0.472 (13) & 0.302 (\phantom{1}8) \\
CometKiwi* & 0.333 (11) & 0.345 (10) & 0.490 (\phantom{1}8) & 0.299 (\phantom{1}9) \\
YiSi-1 & 0.309 (14) & 0.307 (13) & 0.458 (16) & 0.298 (10) \\
BLEURT-20 & 0.349 (\phantom{1}9) & 0.368 (\phantom{1}7) & 0.484 (11) & 0.297 (11) \\
BLCOM\_1 & 0.374 (\phantom{1}6) & 0.327 (12) & 0.488 (\phantom{1}9) & 0.295 (12) \\
MEE4 & 0.312 (12) & 0.240 (21) & 0.446 (19) & 0.280 (13) \\
PrismRefMedium & 0.351 (\phantom{1}8) & 0.267 (17) & 0.462 (15) & 0.279 (14) \\
BERTScore & 0.282 (15) & 0.292 (15) & 0.451 (18) & 0.268 (15) \\
PrismRefSmall & 0.339 (10) & 0.276 (16) & 0.457 (17) & 0.260 (16) \\
chrfS & 0.275 (17) & 0.237 (22) & 0.444 (20) & 0.256 (17) \\
XCOMET-QE* & 0.310 (13) & 0.367 (\phantom{1}8) & 0.463 (14) & 0.251 (18) \\
chrF & 0.211 (19) & 0.192 (25) & 0.436 (23) & 0.146 (19) \\
spBLEU & 0.200 (20) & 0.218 (24) & 0.436 (22) & 0.138 (20) \\
\texttt{sentinel-cand-mqm}* & 0.140 (22) & 0.262 (19) & 0.481 (12) & 0.108 (21) \\
bright-qe* & 0.195 (21) & 0.301 (14) & 0.484 (10) & 0.103 (22) \\
XLsimMqm* & 0.056 (24) & 0.224 (23) & 0.438 (21) & 0.082 (23) \\
BLEU & 0.078 (23) & 0.079 (26) & 0.435 (26) & 0.019 (24) \\
\texttt{sentinel-ref-mqm} & 0.000 (25) & 0.263 (18) & 0.435 (24) & 0.000 (25) \\
\texttt{sentinel-src-mqm}* & 0.000 (25) & 0.243 (20) & 0.435 (24) & 0.000 (25) \\
\bottomrule
\end{tabular}
\end{adjustbox}
\caption{\label{ja-zh_metric_comparison_table}
The scores (and ranks) of the metrics as evaluated by Pearson, $acc_{eq}$, and PDP using segment-level correlation on the WMT’24 ja$\rightarrow$zh dataset, sorted by PDP rank. QE metrics are marked with a *.}
\end{table}
\begin{table}[]
\begin{adjustbox}{width=\columnwidth}
\begin{tabular}{lrrrr}
\toprule
\textbf{Metric} & \textbf{Segment-Wise Pearson} & \textbf{Global Pearson} & $\mathbf{acc_{eq}}$ & $\mathbf{PDP}$\\
\midrule
metametrics & 0.249 (\phantom{1}2) & 0.339 (\phantom{1}1) & 0.686 (\phantom{1}4) & 0.285 (\phantom{1}1) \\
XCOMET & 0.241 (\phantom{1}4) & 0.331 (\phantom{1}2) & 0.688 (\phantom{1}2) & 0.285 (\phantom{1}2) \\
MetricX-24-Hybrid & 0.241 (\phantom{1}3) & 0.326 (\phantom{1}3) & 0.685 (\phantom{1}6) & 0.275 (\phantom{1}3) \\
MetricX-24-Hybrid-QE* & 0.229 (\phantom{1}5) & 0.299 (\phantom{1}6) & 0.685 (\phantom{1}7) & 0.264 (\phantom{1}4) \\
XCOMET-QE* & 0.204 (\phantom{1}8) & 0.308 (\phantom{1}4) & 0.687 (\phantom{1}3) & 0.254 (\phantom{1}5) \\
bright-qe* & 0.160 (14) & 0.302 (\phantom{1}5) & 0.689 (\phantom{1}1) & 0.249 (\phantom{1}6) \\
\texttt{sentinel-cand-mqm}* & 0.198 (11) & 0.264 (\phantom{1}8) & 0.683 (10) & 0.229 (\phantom{1}7) \\
gemba\_esa* & 0.221 (\phantom{1}7) & 0.252 (11) & 0.683 (11) & 0.227 (\phantom{1}8) \\
COMET-22 & 0.265 (\phantom{1}1) & 0.257 (\phantom{1}9) & 0.683 (12) & 0.227 (\phantom{1}9) \\
metametrics\_qe* & 0.153 (16) & 0.286 (\phantom{1}7) & 0.686 (\phantom{1}5) & 0.207 (10) \\
CometKiwi* & 0.201 (10) & 0.214 (13) & 0.684 (\phantom{1}8) & 0.205 (11) \\
BLCOM\_1 & 0.228 (\phantom{1}6) & 0.227 (12) & 0.681 (16) & 0.189 (12) \\
BLEURT-20 & 0.203 (\phantom{1}9) & 0.253 (10) & 0.681 (17) & 0.185 (13) \\
BERTScore & 0.183 (12) & 0.179 (17) & 0.682 (13) & 0.143 (14) \\
MEE4 & 0.151 (17) & 0.138 (19) & 0.683 (9) & 0.135 (15) \\
YiSi-1 & 0.179 (13) & 0.157 (18) & 0.681 (18) & 0.129 (16) \\
chrfS & 0.150 (18) & 0.123 (20) & 0.682 (15) & 0.121 (17) \\
damonmonli & 0.088 (23) & 0.194 (15) & 0.682 (14) & 0.100 (18) \\
PrismRefMedium & 0.147 (19) & 0.116 (21) & 0.680 (20) & 0.090 (19) \\
chrF & 0.131 (20) & 0.115 (22) & 0.680 (24) & 0.087 (20) \\
PrismRefSmall & 0.153 (15) & 0.114 (23) & 0.680 (22) & 0.086 (21) \\
spBLEU & 0.121 (21) & 0.113 (24) & 0.680 (21) & 0.067 (22) \\
BLEU & 0.104 (22) & 0.103 (25) & 0.680 (23) & 0.059 (23) \\
\texttt{sentinel-ref-mqm} & 0.000 (25) & 0.180 (16) & 0.680 (25) & 0.000 (24) \\
\texttt{sentinel-src-mqm}* & 0.000 (25) & 0.194 (14) & 0.680 (25) & 0.000 (24) \\
XLsimMqm* & 0.018 (24) & 0.032 (24) & 0.681 (19) & -0.011 (26) \\
\bottomrule
\end{tabular}
\end{adjustbox}
\caption{\label{en-es_metric_comparison_table}
The scores (and ranks) of the metrics as evaluated by Pearson, $acc_{eq}$, and PDP using segment-level correlation on the WMT’24 en$\rightarrow$es dataset, sorted by PDP rank. QE metrics are marked with a *.}
\end{table}

\begin{table}[]
\begin{adjustbox}{width=\columnwidth}
\begin{tabular}{lrrrr}
\toprule
\textbf{Metric} & \textbf{Segment-Wise Pearson} & \textbf{Global Pearson} & $\mathbf{acc_{eq}}$ & $\mathbf{PDP}$\\
\midrule
human-round2 & 0.490 (\phantom{1}9) & 0.656 (\phantom{1}4) & 0.586 (\phantom{1}5) & 0.568 (\phantom{1}1) \\
human-round3 & 0.459 (15) & 0.722 (\phantom{1}1) & 0.577 (\phantom{1}8) & 0.548 (\phantom{1}2) \\
MetricX-23-QE* & 0.511 (\phantom{1}3) & 0.626 (\phantom{1}5) & 0.596 (\phantom{1}3) & 0.501 (\phantom{1}3) \\
XCOMET-Ensemble & 0.538 (\phantom{1}2) & 0.695 (\phantom{1}2) & 0.604 (\phantom{1}1) & 0.488 (\phantom{1}4) \\
MetricX-23 & 0.507 (\phantom{1}5) & 0.585 (\phantom{1}6) & 0.603 (\phantom{1}2) & 0.474 (\phantom{1}5) \\
XCOMET-QE-Ensemble* & 0.507 (\phantom{1}6) & 0.679 (\phantom{1}3) & 0.588 (\phantom{1}4) & 0.463 (\phantom{1}6) \\
COMET & 0.508 (\phantom{1}4) & 0.432 (18) & 0.574 (\phantom{1}9) & 0.449 (\phantom{1}7) \\
docWMT22CometDA & 0.484 (10) & 0.394 (21) & 0.559 (14) & 0.446 (\phantom{1}8) \\
cometoid22-wmt22* & 0.499 (\phantom{1}7) & 0.441 (16) & 0.578 (\phantom{1}7) & 0.401 (\phantom{1}9) \\
GEMBA-MQM* & 0.482 (11) & 0.502 (11) & 0.572 (11) & 0.399 (10) \\
BLEURT-20 & 0.492 (\phantom{1}8) & 0.484 (12) & 0.572 (10) & 0.389 (11) \\
Calibri-COMET22 & 0.477 (12) & 0.413 (20) & 0.522 (25) & 0.380 (12) \\
Yisi-1 & 0.404 (19) & 0.366 (22) & 0.542 (18) & 0.372 (13) \\
sescoreX & 0.459 (14) & 0.519 (\phantom{1}9) & 0.563 (13) & 0.359 (14) \\
mbr-metricx-qe* & 0.543 (\phantom{1}1) & 0.571 (\phantom{1}7) & 0.584 (\phantom{1}6) & 0.345 (15) \\
CometKiwi* & 0.463 (13) & 0.475 (13) & 0.569 (12) & 0.341 (16) \\
KG-BERTScore* & 0.456 (16) & 0.451 (14) & 0.556 (15) & 0.339 (17) \\
BERTScore & 0.355 (24) & 0.325 (23) & 0.528 (22) & 0.336 (18) \\
docWMT22CometKiwiDA* & 0.426 (18) & 0.444 (15) & 0.547 (16) & 0.334 (19) \\
MaTeSe & 0.330 (29) & 0.554 (\phantom{1}8) & 0.528 (21) & 0.324 (20) \\
XLSim & 0.372 (22) & 0.239 (26) & 0.527 (23) & 0.320 (21) \\
Calibri-COMET22-QE* & 0.432 (17) & 0.441 (17) & 0.483 (32) & 0.318 (22) \\
MS-COMET-QE-22* & 0.400 (20) & 0.310 (24) & 0.546 (17) & 0.306 (23) \\
tokengram\_F & 0.340 (27) & 0.227 (29) & 0.520 (26) & 0.301 (24) \\
chrF & 0.336 (28) & 0.232 (28) & 0.519 (28) & 0.300 (25) \\
f200spBLEU & 0.343 (26) & 0.237 (27) & 0.526 (24) & 0.274 (26) \\
embed\_llama & 0.242 (32) & 0.250 (25) & 0.483 (31) & 0.254 (27) \\
MEE4 & 0.360 (23) & 0.202 (30) & 0.529 (20) & 0.250 (28) \\
BLEU & 0.310 (31) & 0.192 (31) & 0.520 (27) & 0.242 (29) \\
mre-score-labse-regular & 0.376 (21) & 0.111 (32) & 0.530 (19) & 0.208 (30) \\
prismRef & 0.349 (25) & 0.516 (10) & 0.518 (29) & 0.121 (31) \\
random-sysname* & 0.124 (33) & 0.064 (33) & 0.409 (34) & 0.114 (32) \\
eBLEU & 0.317 (30) & -0.011 (34) & 0.512 (30) & 0.094 (33) \\
prismSrc* & 0.102 (34) & 0.425 (19) & 0.426 (33) & -0.139 (34) \\
\bottomrule
\end{tabular}
\end{adjustbox}
\caption{\label{wmt23_en-de_metric_comparison_table}
Scores (and ranks) of metrics evaluated by Segment-Wise Pearson, Global Pearson, $acc_{eq}$, and PDP on the WMT’23 en$\rightarrow$de dataset. QE metrics are marked with a *.}
\end{table}
\begin{table}[]
\begin{adjustbox}{width=\columnwidth}
\begin{tabular}{lrrrr}
\toprule
\textbf{Metric} & \textbf{Segment-Wise Pearson} & \textbf{Global Pearson} & $\mathbf{acc_{eq}}$ & $\mathbf{PDP}$\\
\midrule
XCOMET-Ensemble & 0.421 (\phantom{1}3) & 0.650 (\phantom{1}1) & 0.543 (\phantom{1}1) & 0.477 (\phantom{1}1) \\
human-round3 & 0.393 (\phantom{1}5) & 0.611 (\phantom{1}5) & 0.522 (\phantom{1}8) & 0.463 (\phantom{1}2) \\
XCOMET-QE-Ensemble* & 0.380 (\phantom{1}7) & 0.647 (\phantom{1}3) & 0.533 (\phantom{1}3) & 0.449 (\phantom{1}3) \\
MetricX-23-QE* & 0.359 (12) & 0.647 (\phantom{1}2) & 0.527 (\phantom{1}5) & 0.442 (\phantom{1}4) \\
human-round2 & 0.403 (\phantom{1}4) & 0.572 (\phantom{1}6) & 0.523 (\phantom{1}7) & 0.431 (\phantom{1}5) \\
mbr-metricx-qe* & 0.436 (\phantom{1}1) & 0.489 (\phantom{1}9) & 0.537 (\phantom{1}2) & 0.431 (\phantom{1}6) \\
MetricX-23 & 0.373 (\phantom{1}8) & 0.625 (\phantom{1}4) & 0.531 (\phantom{1}4) & 0.428 (\phantom{1}7) \\
GEMBA-MQM* & 0.434 (\phantom{1}2) & 0.449 (11) & 0.522 (\phantom{1}9) & 0.408 (\phantom{1}8) \\
CometKiwi* & 0.388 (\phantom{1}6) & 0.442 (13) & 0.525 (\phantom{1}6) & 0.399 (\phantom{1}9) \\
KG-BERTScore* & 0.369 (10) & 0.430 (14) & 0.516 (11) & 0.392 (10) \\
MaTeSe & 0.325 (19) & 0.511 (\phantom{1}8) & 0.479 (26) & 0.362 (11) \\
docWMT22CometKiwiDA* & 0.340 (15) & 0.387 (17) & 0.493 (19) & 0.360 (12) \\
cometoid22-wmt22* & 0.357 (13) & 0.479 (10) & 0.515 (12) & 0.352 (13) \\
Calibri-COMET22-QE* & 0.355 (14) & 0.443 (12) & 0.491 (21) & 0.348 (14) \\
BLEURT-20 & 0.371 (\phantom{1}9) & 0.378 (18) & 0.518 (10) & 0.347 (15) \\
COMET & 0.364 (11) & 0.396 (15) & 0.514 (13) & 0.345 (16) \\
MS-COMET-QE-22* & 0.306 (22) & 0.367 (19) & 0.498 (18) & 0.324 (17) \\
docWMT22CometDA & 0.327 (18) & 0.353 (20) & 0.493 (20) & 0.324 (18) \\
Yisi-1 & 0.329 (17) & 0.290 (21) & 0.504 (14) & 0.321 (19) \\
sescoreX & 0.295 (23) & 0.536 (\phantom{1}7) & 0.499 (16) & 0.303 (20) \\
BERTScore & 0.309 (21) & 0.236 (22) & 0.499 (17) & 0.294 (21) \\
Calibri-COMET22 & 0.311 (20) & 0.396 (16) & 0.474 (28) & 0.293 (22) \\
prismRef & 0.332 (16) & 0.183 (24) & 0.504 (15) & 0.284 (23) \\
tokengram\_F & 0.262 (25) & 0.060 (32) & 0.485 (23) & 0.218 (24) \\
chrF & 0.263 (24) & 0.063 (31) & 0.485 (22) & 0.212 (25) \\
mre-score-labse-regular & 0.251 (26) & 0.145 (26) & 0.481 (24) & 0.207 (26) \\
XLSim & 0.218 (30) & 0.111 (28) & 0.464 (31) & 0.189 (27) \\
f200spBLEU & 0.220 (28) & 0.108 (29) & 0.476 (27) & 0.169 (28) \\
MEE4 & 0.236 (27) & 0.105 (30) & 0.480 (25) & 0.163 (29) \\
eBLEU & 0.219 (29) & -0.084 (34) & 0.473 (29) & 0.156 (30) \\
BLEU & 0.208 (31) & 0.119 (27) & 0.472 (30) & 0.152 (31) \\
embed\_llama & 0.138 (32) & 0.161 (25) & 0.447 (32) & 0.120 (32) \\
prismSrc* & 0.078 (33) & 0.223 (23) & 0.421 (33) & 0.054 (33) \\
random-sysname* & 0.019 (34) & 0.018 (33) & 0.381 (34) & 0.021 (34) \\
\bottomrule
\end{tabular}
\end{adjustbox}
\caption{\label{wmt23_zh-en_metric_comparison_table}
Scores (and ranks) of metrics evaluated by Segment-Wise Pearson, Global Pearson, $acc_{eq}$, and PDP on the WMT’23 zh$\rightarrow$en dataset. QE metrics are marked with a *.}
\end{table}

\end{document}